\documentclass[final]{cvpr}

\usepackage{times}
\usepackage{epsfig}
\usepackage{graphicx}
\usepackage{amsmath}
\usepackage{amssymb}

\usepackage{bm}
\usepackage{multirow}
\usepackage{multicol}
\usepackage{floatrow}
\usepackage[normalem]{ulem}
\usepackage{graphics}
\usepackage{algpseudocode}
\usepackage{caption}
\usepackage[accsupp]{axessibility} 
\usepackage{makecell}
\usepackage{booktabs}
\usepackage[table]{xcolor}
\definecolor{TableBlue}{HTML}{E0FFFF}

\usepackage[pagebackref,breaklinks,colorlinks]{hyperref}

\usepackage{xcolor}
\definecolor{gray}{rgb}{0.5,0.5,0.5} 
\definecolor{frenchblue}{rgb}{0.0, 0.45, 0.73}
\definecolor{gray}{rgb}{0.5,0.5,0.5} 
\definecolor{green}{rgb}{0, 0.4, 0} 
\definecolor{orange}{rgb}{1, 0.5, 0} 	
\definecolor{mahogany}{rgb}{0.75, 0.25, 0.0}
\definecolor{purple}{rgb}{0.6, 0, 0.6}
\definecolor{darkgreen}{rgb}{0, 0.4, 0.4} 
\definecolor{red}{rgb}{1.0, 0, 0}
\definecolor{plotpurple}{rgb}{0.2353, 0.2, 0.90196}
\definecolor{plotorange}{rgb}{1.0, 0.6, 0.2}
\definecolor{plotgreen}{rgb}{0.2, 0.784313, 0.2}
\definecolor{plotred}{rgb}{1.0, 0.2, 0.392}

\newboolean{revising}
\setboolean{revising}{true}

\newcommand{\red}[1]{\textcolor{red}{\textbf{#1}}}
\newcommand{\blue}[1]{\textcolor{blue}{\textbf{#1}}}

\newcommand{\apBev}{\ap$_{\text{BEV}}$}
\newcommand{\apForty}{\ap$_{40}$}
\newcommand{\apEleven}{\ap$_{11}$}
\newcommand{\ap}{AP}

\newcommand{\apthreeD}{\ap$_{3\text{D}}$}

\def\cvprPaperID{} 
\def\confName{CVPR}
\def\confYear{2022}

\begin{document}

\title{MonoDTR: Monocular 3D Object Detection with Depth-Aware Transformer}

\author{First Author\\
Institution1\\
Institution1 address\\
{\tt\small firstauthor@i1.org}
\and
Second Author\\
Institution2\\
First line of institution2 address\\
{\tt\small secondauthor@i2.org}
}

\author{Kuan-Chih Huang$^1$ \quad Tsung-Han Wu$^1$ \quad Hung-Ting Su$^1$ \quad Winston H. Hsu$^{1,2}$\\
$^1$ National Taiwan University \quad $^2$ Mobile Drive Technology
}
\maketitle

\urlstyle{same}

\pagestyle{empty}  
\thispagestyle{empty}

\begin{abstract}
    Monocular 3D object detection is an important yet challenging task in autonomous driving.
    Some existing methods leverage depth information from an off-the-shelf depth estimator to assist 3D detection, but suffer from the additional computational burden and achieve limited performance caused by inaccurate depth priors.
    To alleviate this, we propose MonoDTR, a novel end-to-end depth-aware transformer network for monocular 3D object detection. It mainly consists of two components: (1) the Depth-Aware Feature Enhancement (DFE) module that implicitly learns depth-aware features with auxiliary supervision without requiring extra computation, and (2) the Depth-Aware Transformer (DTR) module that globally integrates context- and depth-aware features. Moreover, different from conventional pixel-wise positional encodings, we introduce a novel depth positional encoding (DPE) to inject depth positional hints into transformers.
    Our proposed depth-aware modules can be easily plugged into existing image-only monocular 3D object detectors to improve the performance.
    Extensive experiments on the KITTI dataset demonstrate that our approach outperforms previous state-of-the-art monocular-based methods and achieves real-time detection. Code is available at \url{https://github.com/kuanchihhuang/MonoDTR}.
    
\end{abstract}

\section{Introduction}

\begin{figure}
    \centering
    \includegraphics[width=1\linewidth]{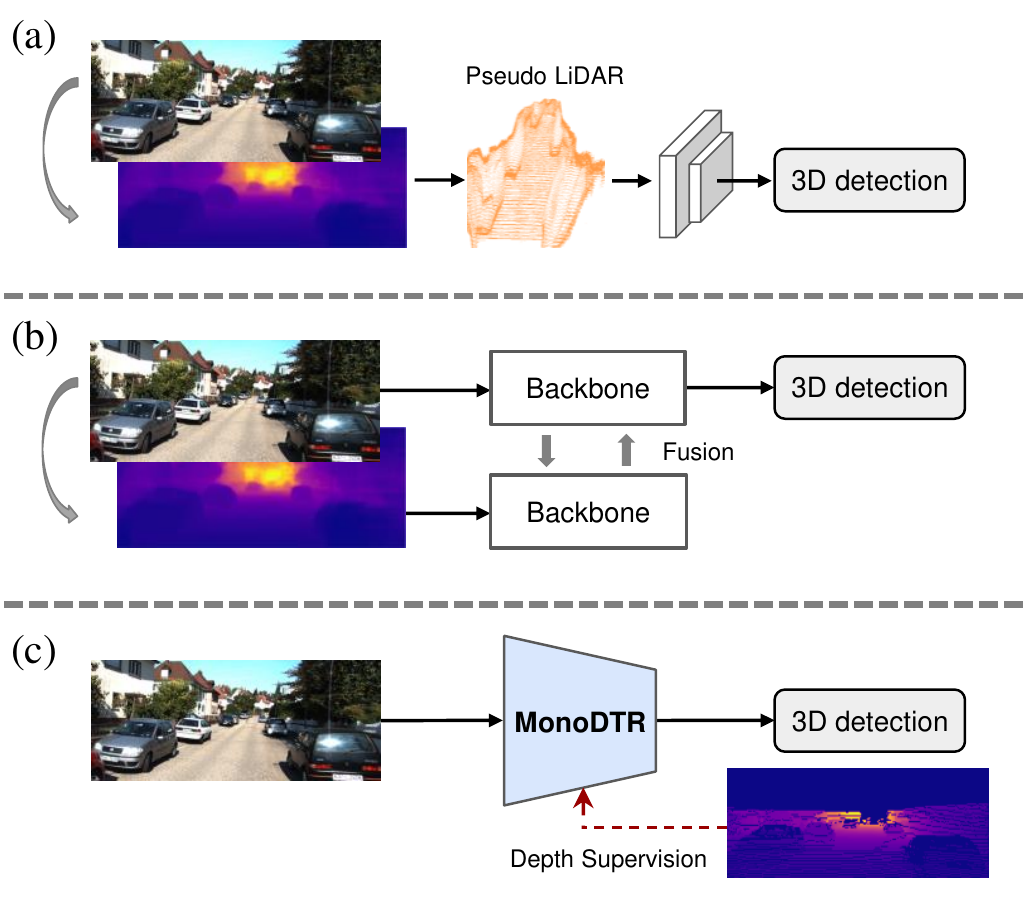} %
    \caption{
    \textbf{
    Comparison of different depth-assisted monocular 3D object detection frameworks.
    } 
    (a) Pseudo-LiDAR-based methods \cite{wang2019pseudo, weng2019mono3dplidar, ma2019am3d} lift images to 3D coordinate via monocular depth estimation, followed by a 3D LiDAR-based detector to recover object locations. (b) Fusion-based methods \cite{ding2020d4lcn, ouyang2021ddf, wang2021ddmp} extract features from images and estimated depth maps, then fuse them to predict objects.
    (c) Our MonoDTR learns depth-aware features via additional depth supervision and performs 3D object detection in an end-to-end manner. Note that our depth supervision is only leveraged in the training stage.
    }
    \label{fig:example}
    \vspace{-12.5pt}
\end{figure}

Three-dimensional (3D) object detection is a fundamental problem and enables various applications such as autonomous driving. Previous methods have achieved superior performance based on the accurate depth information from multiple sensors, such as LiDAR signal \cite{Lang2019pointpillars, shi2019pointrcnn, he2020sassd, shi2020pv} or stereo matching \cite{li2019stereorcnn, wang2019pseudo,chen2020dsgn, sun2020disprcnn}. In order to lower the sensor costs, some image-only monocular 3D object detection methods \cite{Mousavian2017deep3dbox,ku2019monopsr, ma2019am3d, chen2020monopair, brazil2020kinematic,wang2021fcos3d} have been proposed and made impressive progress relying on geometry constraints between 2D and 3D. However, the performance is still far from satisfactory without the aid of depth cues.

Recently, several works have tried to produce estimated depth from the pre-trained depth estimation models to assist monocular 3D object detection.
Pseudo-LiDAR-based approaches \cite{ma2019am3d, wang2019pseudo, weng2019mono3dplidar} convert estimated depth maps into 3D point clouds to imitate LiDAR signals,
followed by the existing LiDAR-based detector for 3D object detection (see Figure \ref{fig:example}(a)). 
Some fusion-based approaches \cite{ding2020d4lcn, wang2021ddmp, ouyang2021ddf} apply several fusion strategies to combine features extracted from depths and images to detect objects (see Figure \ref{fig:example}(b)).
These methods, though better localize objects with the help of estimated depth, may suffer from the risk of learning 3D detection on inaccurate depth maps.
Also, the additional computational cost of the depth estimator makes it impractical for real-world applications \cite{Ma2021monodle}.

To address the above issues, we propose MonoDTR, a novel end-to-end depth-aware transformer network for monocular 3D object detection (see Figure \ref{fig:example}(c)).
A depth-aware feature enhancement (DFE) module is introduced to learn depth-aware features with the auxiliary depth supervision, which avoids obtaining inaccurate depth priors from the pre-trained depth estimator.
Furthermore, the DFE module is lightweight yet effective in assisting 3D object detection without constructing the complicated architecture to extract features from off-the-shelf depth maps. It significantly reduces computational time compared with previous depth-assisted methods \cite{ma2019am3d, ding2020d4lcn, wang2021ddmp} (see Table \ref{tab:kitti_test_car}).

In addition, unlike previous fusion-based methods (\eg, D$^4$LCN\cite{ding2020d4lcn} and DDMP-3D\cite{wang2021ddmp}) that apply carefully designed convolution kernels for context- and depth-aware features, 
we develop the first transformer-based fusion module to globally integrate the image and depth information.
The transformer encoder-decoder structure \cite{vaswani2017SA} has been proven to capture long-range dependency effectively; thus, we apply it to model the relationship between context- and depth-aware features.
To better represent the property of the 3D object, we utilize depth-aware features to replace the commonly used object queries \cite{Nicolas2020detr, zhu2021deform, kim2021hotr} as input of the transformer decoder, which can provide more meaningful cues for 3D reasoning.
Furthermore, we introduce a novel depth positional encoding (DPE) to involve depth-aware hints to the transformer, achieving better performance than conventional pixel-wise positional encodings.

We summarize our contributions as follows:
\begin{enumerate}
    \item
    We propose a novel framework, MonoDTR, learning depth-aware features via auxiliary supervision to assist monocular 3D object detection,
    which avoids introducing high computational cost and inaccurate depth priors from using the off-the-shelf depth estimator.

    \item We present the first depth-aware transformer module to integrate context- and depth-aware features efficiently. A novel depth positional encoding (DPE) is proposed to inject depth positional hints into the transformer. 
    
    \item Experimental results on the KITTI dataset show that our approach outperforms state-of-the-art monocular-based methods and achieves real-time detection. Furthermore, the proposed depth-aware modules can be easily plug-and-play in existing image-only frameworks to improve performance.

\end{enumerate}

\section{Related Work}

\noindent {\bf Image-only monocular 3D object detection.}
Recently, several works only adopt a single image for 3D object detection \cite{Mousavian2017deep3dbox, andrea2019monodis, brazil2019m3drpn, roddick2019oft, liu2020SMOKE, Simonelli2020MoVi3D, liu2021ground, wang2021pgd}. Due to the lack of depth information from images, these methods mainly rely on geometric consistency to predict objects.
Deep3Dbox \cite{Mousavian2017deep3dbox} solves orientation prediction by proposed novel {\it MultiBin} loss and enforces constraint between 2D and 3D boxes with geometric prior. 
M3D-RPN \cite{brazil2019m3drpn} generates 3D object proposals with 2D bounding box constraints and proposes a depth-aware convolution to predict 3D objects.
OFTNet\cite{roddick2018orthographic} introduces an orthographic feature transform to map image-based features into a 3D voxel space.
Besides, MonoPair \cite{chen2020monopair} explores spatial pair-wise relationship between objects to improve detection performance.
M3DSSD \cite{luo2021m3dssd} presents a two-step feature alignment approach to solve the feature mismatching problem.
Furthermore, some works \cite{liu2020SMOKE,li2020rtm3d, Ma2021monodle, Zhang2021MonoFlex} predict keypoints of the 3D bounding box as an intermediate task to recover the location of objects. 
However, the above purely monocular methods fail to accurately localize objects due to the lack of depth cues.

\noindent {\bf Depth-assisted monocular 3D object detection.}
To further improve the performance, many approaches propose using depth information to aid 3D object detection \cite{you2020pseudo, weng2019mono3dplidar, ma2019am3d, ma2020patchnet, ding2020d4lcn, wang2021ddmp}. Some prior works \cite{wang2019pseudo, weng2019mono3dplidar, ma2019am3d} transform image into pseudo-LiDAR representation by leveraging off-the-shelf depth estimator and calibration parameters, followed by the existing LiDAR-based 3D detector to predict objects, leading to progressive improvement. PatchNet \cite{ma2020patchnet} reveals that the success of pseudo-LiDAR comes from the coordinate transformation and organizes it into the image representation, which can benefit from the powerful CNNs networks.
D$^4$LCN\cite{ding2020d4lcn} and DDMP-3D\cite{wang2021ddmp} focus on developing the fusion-based approach between image and estimated depth with carefully designed convolutional networks. 
Besides, CaDDN \cite{Reading2021CaDDN} learns categorical depth distributions for each pixel to construct bird’s-eye-view (BEV) representations and recovers bounding boxes from the BEV projection. 
However, most of the abovementioned methods directly using pre-trained depth estimators suffer from additional computational cost and only achieve limited performance caused by inaccurate depth priors.

\noindent {\bf Transformer.}
Transformer \cite{vaswani2017SA} was firstly introduced in sequential modeling and has considerable improvement in natural language processing (NLP) tasks.
The self-attention mechanism is the core component in the transformer with its capability of capturing the long-range dependencies. Recently, transformer architecture has been successfully leveraged in the computer vision field, such as image classification \cite{dosovitskiy2020vit} and human-object interaction \cite{kim2021hotr}. 
In addition, DETR \cite{Nicolas2020detr} proposes developing object detection with the transformer without relying on many hand-designed components used in traditional pipelines.

Though the transformer can perform well in most visual tasks, its usage in monocular 3D object detection has not been explored.
In the image-based 3D detection task, the object size at far and near distance in the image varies significantly due to the perspective projection \cite{ding2020d4lcn, wang2021ddmp}, which makes it challenging to utilize the learned object query mentioned in DETR\cite{Nicolas2020detr} to fully represent the object property.
Thus, in this paper, we propose to globally integrate context- and depth-aware features with transformers and inject depth hints into the transformer for better 3D reasoning.

\begin{figure*}[t]
\centering
\includegraphics[width=1\textwidth]{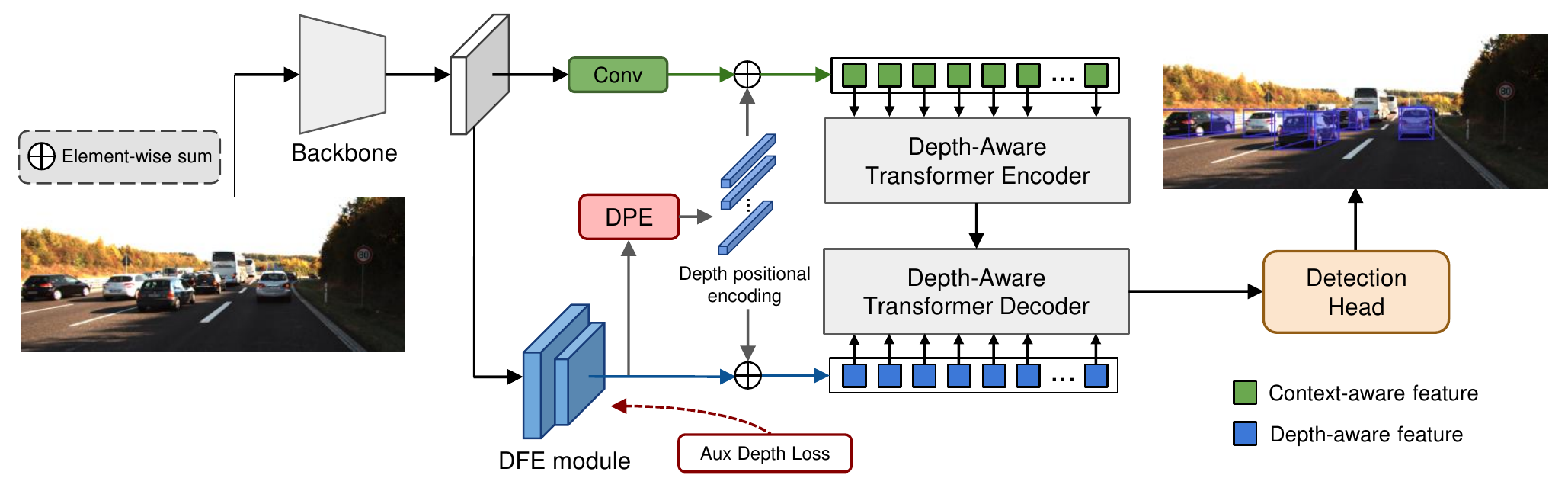}
\caption{\textbf{The overall framework of our proposed MonoDTR.}
The input image is first sent to the backbone to extract the features.
The Depth-Aware Feature Enhancement (DFE) module learns depth-aware features via auxiliary supervision (Section \ref{sec:dam}), and context-aware features are extracted by convolution layers in parallel.
The Depth-Aware Transformer (DTR) module then integrates two kinds of features, while the Depth Positional Encoding (DPE) module injects depth positional hints into the transformer (Section \ref{sec:dgtr}).
Finally, the detection head is applied to predict the 3D bounding boxes (Section \ref{sec:loss}). Note that the auxiliary depth supervision is only used in the training phase.
}
\label{fig:arch}
\vspace{-9.7pt}
\end{figure*}

\section{Proposed Approach}
\label{sec:proposed}
\subsection{Framework Overview}
Figure \ref{fig:arch} presents the framework of our MonoDTR, which mainly consists of four components: the backbone, the depth-aware feature enhancement (DFE) module, the depth-aware transformer (DTR) module, and the 2D-3D detection head. We adopt DLA-102 \cite{fisher2018dla} as our backbone network following \cite{luo2021m3dssd}. Given an input RGB image with resolution $H_{\mathrm{inp}}$ $\times$ $W_{\mathrm{inp}}$, the backbone outputs a feature map $\mathbf{F} \in \mathbb{R}^{C \times H \times W}$, where $H = \frac{H_{\mathrm{inp}}}{8}$, $W =\frac{W_{\mathrm{inp}}}{8}$, and $C=256$. 
The DFE module is presented to implicitly learn depth-aware features (Section \ref{sec:dam}), while several convolution layers are applied to extract context-aware features in parallel.
Then, we globally integrate two kinds of features by the DTR module and first attempt to inject depth positional hints into the transformer through the depth positional encoding (DPE) module (Section \ref{sec:dgtr}).
Consequently, the anchor-based detection heads and loss functions are adopted for 2D and 3D object detection (Section \ref{sec:loss}).

\subsection{Depth-Aware Feature Enhancement Module}
\label{sec:dam}

\begin{figure}[thpb]
    \centering
    \includegraphics[width=\columnwidth]{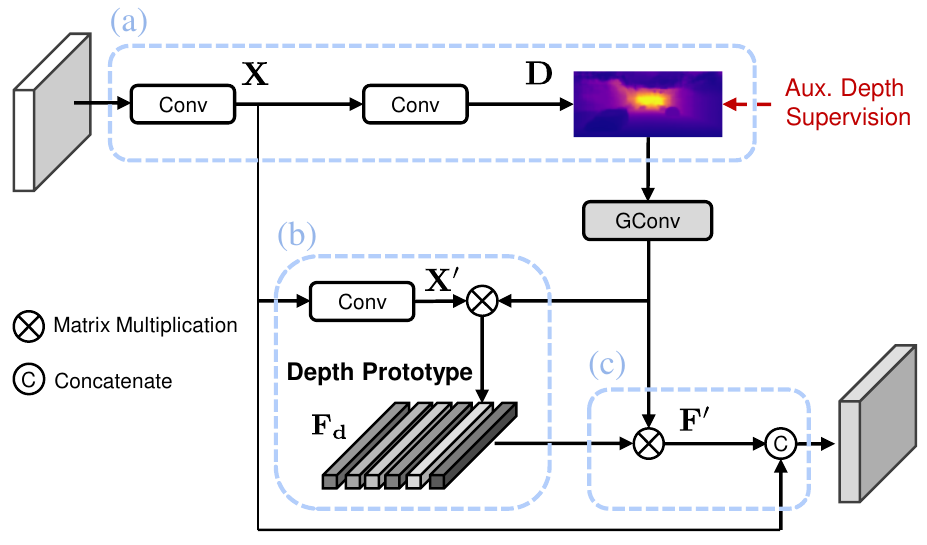}

    \vspace{-8pt}
    \caption{\textbf{The architecture of depth-aware feature enhancement (DFE) module.}
    The DFE module aims to implicitly learn depth-aware features via auxiliary supervision.
    (a) Generate initial depth-aware feature $\mathbf{X}$ and predict depth distribution $\mathbf{D}$. (b) Estimate feature representation of depth prototype $\mathbf{F_d}$. (c) Produce depth prototype enhanced feature $\mathbf{F'}$, and fuse with initial depth-aware feature $\mathbf{X}$. See Section \ref{sec:dam} for details.
    }
    \label{fig:dam}
\end{figure}

Existing depth-assisted methods \cite{ding2020d4lcn, wang2021ddmp, wang2019pseudo}, using off-the-shelf depth estimators, suffer from the risk of introducing inaccurate depth priors and extra computation burden.
To alleviate this, we propose a depth-aware feature enhancement (DFE) module for depth reasoning as in Figure \ref{fig:dam}.
The precise depth map is utilized for auxiliary supervision in the training stage, making the DFE module implicitly learn the depth-aware features.
Compared with previous works that apply an additional backbone \cite{ding2020d4lcn, wang2021ddmp} or complicated architectures \cite{Reading2021CaDDN} to encode depths, we generate depth-aware features to assist 3D object detection with a lightweight module, significantly reducing the computation budget.

\smallskip\noindent\textbf{Learning initial depth-aware feature.}
To generate depth-aware features, we leverage an auxiliary depth estimation task and consider it as a sequential classification problem \cite{FuCVPR18-DORN, Reading2021CaDDN}.
As illustrated in Figure \ref{fig:dam}(a), given the input feature $\mathbf{F} \in \mathbb{R}^{C \times H \times W}$ from the backbone, we adopt two convolution layers to predict the probability of discretized depth bins $\mathbf{D} \in \mathbb{R}^{D \times H \times W}$, where $D$ is the number of depth categories (bins). 
The probability represents the confidence that the depth value of each pixel belongs to a certain depth bin.
To discretize the depth ground truth from continuous space to discretization intervals, we utilize linear-increasing discretization (LID) \cite{tang2020center3d, Reading2021CaDDN} to formulate the depth bins (more details can be found in supplementary materials).
To this end, the intermediate feature map $\mathbf{X} \in \mathbb{R}^{C \times H \times W}$ can be regarded as initial depth-aware features.

\smallskip\noindent\textbf{Depth prototype representation learning.}
To further enhance the capability of depth representation, we augment the feature of each pixel by introducing the central representation of the corresponding depth category (bin), inspired from the class center in \cite{zhang2019acfnet}.
The feature center of each depth category (regarded as the depth prototype) can be computed by aggregating the depth-aware features of each pixel belonging to a specified category.
In practice, we first apply a group convolution \cite{Krizhevsky2012alexnet} to the predicted depth map $\mathbf{D}$ to merge the adjacent depth categories (bins), reducing the class number from $D$ to $D'=D/r$ with scale $r$. It helps to share similar depth cues and reduce computation.
The representation of depth prototype $\mathbf{F}_d$ can be generated by gathering the feature of all pixels $\mathbf{X'}$ weighted by their probability to the depth category $d$: 
\begin{align}
    \mathbf{F}_d = \sum_{i \in \mathcal{I}} {\tilde{P}_{di}} \mathbf{X'}_i,  d=\{1,..., D'\},
    \label{eq:regionrepresentation}
\end{align}
where $\mathbf{X'}_i$ denotes the feature of the $i$th pixel in $\mathbf{X'}$, $\mathcal{I} \in \mathbb{R}^{H \times W}$ is the set of pixels in the feature map, and $\tilde{P}_{di}$ is the normalized probability to $d$th depth prototype. In this manner, $\mathbf{F}_d$ can express the global context information of each depth category as shown in Figure \ref{fig:dam}(b). 

\smallskip\noindent\textbf{Feature enhancement with depth prototype.}
Now we can reconstruct new depth-aware features based on the depth prototype representation, 
which allows each pixel to understand the presentation of the depth category from the global view.
The reconstructed feature $\mathbf{F}'$ is calculated as:
\begin{align}
    \mathbf{F}' = \sum_{d=1}^{D'} {\tilde{P}_{di}} \mathbf{F}_d. \label{eq:recons}
\end{align}
Consequently, we obtain the enhanced depth feature by concatenating the initial depth-aware feature $\mathbf{X}$ and the reconstructed features $\mathbf{F}'$, followed by a simple 1 $\times$ 1 convolution layer, as shown in Figure \ref{fig:dam}(c).

\subsection{Depth-Aware Transformer}
\label{sec:dgtr}

Inspired by the tremendous success of the transformer \cite{vaswani2017SA} on modeling the long-range relationships, we exploit the transformer encoder-decoder architecture to construct the depth-aware transformer (DTR) module to globally integrate the context- and depth-aware features.

\smallskip\noindent\textbf{Transformer encoder.}
Our transformer encoder aims to improve context-aware features, which is built similar to previous works \cite{Nicolas2020detr, zou2021_hoitrans}. The main component in the transformer is the self-attention mechanism \cite{vaswani2017SA}. 
Given the inputs: query $\mathbf{Q} \in \mathbb{R}^{N \times C}$, key $\mathbf{K} \in \mathbb{R}^{N \times C}$, and value $\mathbf{V} \in \mathbb{R}^{N \times C}$ 
with sequence length $N$, 
a single head self-attention layer can be briefly formulated as:
\begin{equation}
    {\rm{Attention}}(\mathbf{Q},\mathbf{K},\mathbf{V})
    = {\rm{softmax}}(\frac{\mathbf{Q}\mathbf{K}^\top}{\sqrt{C}})\mathbf{V}.
\label{eq-att}
\end{equation}
We take the flatten context-aware feature $ \mathbf{X}_c \in \mathbb{R}^{N \times C}$, where $N=H \times W$, as the input to feed into the transformer encoder. The encoded context-aware feature can be obtained through multi-head self-attention operation and the feed-forward network (FFN). 

\smallskip\noindent\textbf{Transformer decoder.}
The decoder is also built upon the standard transformer architecture.
We propose utilizing the depth-aware features as the input of the decoder instead of learnable embeddings (object query) \cite{Nicolas2020detr}, which is different from the common usage in previous encoder-decoder vision transformer works \cite{Nicolas2020detr,zhu2021deform, kim2021hotr, tan2021planeTR}.
The main reason is that, in the monocular 3D object detection task, the camera views at near and far distances often cause significant changes in object scale due to the perspective projection \cite{ding2020d4lcn, wang2021ddmp}.
It makes the simple learnable embedding hard to fully represent the object's property and handle complex scale variant situations. In contrast, plentiful distance-aware cues are hidden in the depth-aware features. 
Thus, we propose adopting depth-aware features as the input of the transformer decoder.
To this end, the decoder can take the power of cross-attention modules in the transformer to efficiently model the relationship between context- and depth-aware features, leading to better performance.

\begin{figure}[t]
    \centering
    \includegraphics[width=1\columnwidth]{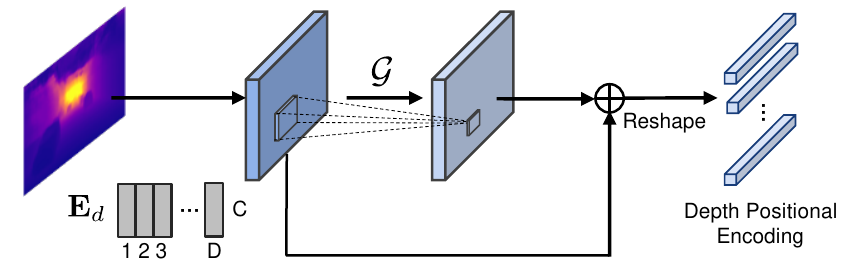}  
    \caption{\textbf{The proposed depth positional encoding (DPE) module.} The DPE module generates depth positional encoding based on the depth category predicted by the DFE module. See Section \ref{sec:dgtr} for details.}
    \label{fig:transformer}
    \vspace{-3mm}
\end{figure}

\smallskip\noindent\textbf{Depth positional encoding (DPE).}
Positional encoding \cite{vaswani2017SA} plays an important role for the transformer to introduce the location information. It is often generated with sinusoidal functions or in a learnable manner according to the pixel location of the image in vision tasks.
Observing that the depth information is much better for the machine to understand the 3D world than the pixel-level relation, we first propose a general depth positional encoding (DPE) module to embed the depth positional hints for each pixel to the transformer.
Specifically, as shown in Figure \ref{fig:transformer}, the depth bin encodings  $\mathbf{E}_{d} = [\bm{e}_{1},\dots,\bm{e}_{D}] \in \mathbb{R}^{D \times C}$
are constructed with learnable embeddings for each depth interval introduced in Section \ref{sec:dam}.
The initial depth positional encoding $\mathbf{P} \in \mathbb{R}^{H \times W \times C}$ can be looked up from $\mathbf{E}_{d}$ according to the argmax of each pixel predicted depth category $\mathbf{D}$.
To further represent the positional cues from local neighborhoods, 
a convolution layer $\mathcal{G}$ with the kernel size of 3 $\times$ 3 is applied and added to $\mathbf{P}$
to obtain final encoding, referred to as depth positional encoding (DPE).

\smallskip\noindent\textbf{Computation reduction.}
The standard self-attention layer in Equation \ref{eq-att} leads to $\mathcal{O}(N^2)$ time and memory, which damages the computational budget.
To mitigate this issue, more recent works \cite{choromanski2020rethinking, katharopoulos2020rnn, wang2020linformer} make efforts on accelerating the attention operation. Among these methods, Linear transformer \cite{katharopoulos2020rnn} proposes to approximate softmax operation with the linear dot product of features. 
Specifically, the similarity function in original transformer \cite{vaswani2017SA} can be formulated as: ${\rm sim}(q,k)={\rm exp}(\frac{q^\top k}{\sqrt{C}})$. 
It is replaced with ${\rm sim}(q,k)=\phi(q)\phi(k)$ in \cite{katharopoulos2020rnn}
, where $\phi(x)=\text{elu}(x)+1$, and $\text{elu}(\cdot)$ is the exponential linear unit
\cite{Clevert2016elu} activation function. 
To this end, $\phi(K)^\top$ and $V$ can be combined first to reduce computation to $\mathcal{O}(N)$. We refer the readers to \cite{katharopoulos2020rnn} for more details. 
In our transformer, we consider applying the linear attention described in \cite{katharopoulos2020rnn} to replace the vanilla self-attention for higher inference speed.

\subsection{2D-3D Detection and Loss }
\label{sec:loss}
\noindent {\bf Anchor definition.}
We adopt the single-stage detector \cite{Joseph2018yolov3, liu2016ssd} with the pre-defined 2D-3D anchors to regress the bounding box.
Each pre-defined anchor consists parameters with 2D bounding box $[x_{2d}, y_{2d}, w_{2d}, h_{2d}]$ and 3D bounding box $[x_{p}, y_{p}, z, w_{3d}, h_{3d}, l_{3d}, \theta]$. $[x_{2d}, y_{2d}]$ and $[x_{p}, y_{p}]$ represent the 2D box center and 3D object center projected to image plane. $[w_{2d}, h_{2d}]$ and ${[w_{3d}, h_{3d}, l_{3d}]}$ represent the physical dimension of 2D and 3D bounding box, respectively. $z$ denotes the depth of 3D object center. $\theta$ is the observation angle.
During training, we project all ground truth into the 2D space to calculate the intersection over union (IoU) with all 2D anchors. 
The anchor with IoU greater than 0.5 is chosen to assign with the corresponding 3D box for optimization.

\smallskip\noindent{\bf Output transformation.}
Similar to prior works \cite{ding2020d4lcn, wang2021ddmp, liu2021ground, luo2021m3dssd}, 
we follow Yolov3 \cite{Joseph2018yolov3} to predict $[t_x, t_y, t_w, t_h]_{2d}$ and $[t_x, t_y, t_w, t_h, t_l, t_z, t_{\theta}]_{3d}$ for each anchor, which aims at parameterizing the residual value for 2D and 3D bounding box, and also predict the classification scores $cls$.
The output bounding box can be restored based on the anchor and the network prediction as follows:
\begin{align}
& [\hat x_{2d},\hat y_{2d}] = [t_x, t_y]_{2d} * [w_{2d}, h_{2d}] + [x_{2d}, y_{2d}] \notag \\
& [\hat x_{p}, \hat x_{p}] = [t_x, t_y]_{3d} * [w_{2d}, h_{2d}] + [x_{p}, y_{p}] \notag \\
& [\hat w_{3d}, \hat h_{3d}, \hat l_{3d}] = \exp([t_w, t_h, t_l]_{3d}) * [w_{3d}, h_{3d}, l_{3d}] \notag \\
& [\hat w_{2d}, \hat h_{2d}] = \exp([t_w, t_h]_{2d}) * [w_{2d}, h_{2d}] \notag \\
& [\hat z, \hat\theta] = [t_z, t_\theta]_{3d} + [z, \theta],
\label{eq:transform}
\end{align}
where $\hat{(\cdot)}$ denotes the recovered parameters of the 3D object. 
Note that we apply the same anchor center for 2D box center $[x_{2d}, y_{2d}]$ and 3D projection center $[x_{p}, y_{p}]$.

\smallskip\noindent {\bf Loss function.}
The overall loss $\mathcal{L}$ contains a classification loss $\mathcal{L}_{cls}$ for objectness and class, a bounding box regression loss $\mathcal{L}_{reg}$ to optimize Equation \ref{eq:transform}, and a depth loss $\mathcal{L}_{dep}$ with auxiliary depth supervision described in Section \ref{sec:dam}:
\vspace{-4mm}
\begin{align}
\mathcal{L} = \mathcal{L}_{cls} + \mathcal{L}_{reg} + \mathcal{L}_{dep}.
\label{eq:loss}
\end{align}
We adopt the focal loss \cite{lin2017focal} to balance the samples for the classification task, and the smoothed-L1 loss \cite{girshickICCV15fastrcnn} for the regression task.
For the depth categorical prediction described in Section \ref{sec:dam}, we utilize the focal loss \cite{lin2017focal}:
\begin{align}
    \mathcal{L}_{dep}=\frac{1}{|\mathcal{P}|}\sum_{p \in \mathcal{P}}\mathrm{FL}(\mathbf{D}(p), \mathbf{\hat{D}}(p)),
\end{align}
where $\mathcal{P}$ is the pixel region on the image with the valid depth labels, and $\mathbf{\hat{D}}$ is the depth bins ground truth generated from LiDAR (more details are provided in the supplementary material).
\vspace{2mm}
\section{Experiments}
\subsection{Setup}
\label{sec:setup}
\noindent{\bf {Dataset.}}
We evaluate the proposed approach on the challenging KITTI 3D object detection dataset \cite{Geiger2012kitti}, which is the most commonly used benchmark for the 3D object detection task. It contains 7481 images for training and 7518 images for testing. We follow \cite{chen2015_3dop} to divide training samples into the training set (3712) and the validation set (3769). The ablation studies are conducted based on this split.

\smallskip\noindent{\bf Evaluation metric.}
The average precision (AP) is used as the metric for evaluation in both 3D object detection and bird's eye view (BEV) detection tasks. We utilize 40 recall positions metric \apForty~instead of original \apEleven~to avoid the bias \cite{andrea2019monodis}. The difficulty of the detection in the benchmark is divided into three levels ("Easy", "Mod.", "Hard") according to size, occlusion, and truncation. 
All methods are ranked based on \apthreeD~of moderate setting (Mod.) same as the KITTI benchmark.
The thresholds of Intersection over Union (IoU) are 0.7, 0.5, 0.5 for cars, cyclists, and pedestrians categories following the official setting.

\begin{table*}[ht]
    \centering
    \small
    \begin{tabular}{l|c|c|ccc|ccc}  
        \hline 
        \multicolumn{1}{l|}{\multirow{2}{*}{Method}} &
        \multicolumn{1}{c|}{\multirow{2}{*}{\hspace{0.5em} Reference \hspace{0.5em}}} &
        \multicolumn{1}{c|}{\multirow{2}{*}{Time(ms)}} & \multicolumn{3}{c|}{\apthreeD@IoU=0.7} &  \multicolumn{3}{c}{\apBev@IoU=0.7} \\ \cline{4-9}  & & & Easy  & Mod. & Hard & Easy & Mod.  & Hard   \\ \hline \hline
        MonoPSR \cite{ku2019monopsr}          & CVPR 2019 & 200   & 10.76 &  7.25 &  5.85 & 18.33 & 12.58 &  9.91 \\
        M3D-RPN \cite{brazil2019m3drpn}       & ICCV 2019 & 160   & 14.76 &  9.71 &  7.42 & 21.02 & 13.67 & 10.23 \\
        MonoPair \cite{chen2020monopair}      & CVPR 2020 & 60    & 13.04 &  9.99 &  8.65 & 19.28 & 14.83 & 12.89 \\
        AM3D \cite{ma2019am3d}                & ICCV 2019 & 400   & 16.50 & 10.74 &  9.52 & 25.03 & 17.32 & 14.91 \\
        MoVi-3D \cite{Simonelli2020MoVi3D}    & ECCV 2020 & 45 & 15.19 & 10.90 & 9.26 & 22.76 & 17.03 & 14.85\\
        PatchNet \cite{ma2020patchnet}        & ECCV 2020 & 400   & 15.68 & 11.12 & 10.17 & 22.97 & 16.86 & 14.97 \\
        M3DSSD \cite{luo2021m3dssd}$\dagger$  & CVPR 2021 &  -    & 17.51 & 11.46 & 8.98  & 24.15 & 15.93 & 12.11 \\
        D4LCN \cite{ding2020d4lcn}            & CVPR 2020 & 200   & 16.65 & 11.72 &  9.51 & 22.51 & 16.02 & 12.55 \\
        MonoDLE \cite{Ma2021monodle}          & CVPR 2021 & 40    & 17.23 & 12.26 & 10.29 & 24.79 & 18.89 & 16.00 \\
        MonoRUn \cite{chen2021monorun}        & CVPR 2021 & 70    & 19.65 & 12.30 & 10.58 & 27.94 & 17.34 & 15.24 \\ 
        GrooMeD-NMS \cite{kumar2021groomed}   & CVPR 2021 & 120   & 18.10 & 12.32 &  9.65 & 26.19 & 18.27 & 14.05 \\
        MonoRCNN \cite{Shi2021MonoRCNN}       & ICCV 2021 & 70    & 18.36 & 12.65 & 10.03 & 25.48 & 18.11 & 14.10 \\ 
        Kinematic3D \cite{brazil2020kinematic}& ECCV 2020 & 120   & 19.07 & 12.72 &  9.17 & 26.69 & 17.52 & 13.10 \\
        DDMP-3D \cite{wang2021ddmp}           & CVPR 2021 & 180   & 19.71 & 12.78 & 9.80  & 28.08 & 17.89 & 13.44 \\ 
        CaDDN \cite{Reading2021CaDDN}         & CVPR 2021 & 630   & 19.17 & 13.41 & 11.46 & 27.94 & 18.91 & \blue{17.19} \\
        DFRNet\cite{Zou2021dfr}        & ICCV 2021 & 180   & 19.40 & 13.63 & 10.35 & 28.17 & 19.17 & 14.84 \\
        MonoEF \cite{zhou2021monoef}          & CVPR 2021 & 30    & \blue{21.29} & 13.87 & 11.71 & \red{29.03} & 19.70 & \red{17.26} \\
        MonoFlex \cite{Zhang2021MonoFlex}     & CVPR 2021 & 30    & 19.94 & 13.89 & \blue{12.07} & 28.23 & \blue{19.75} & 16.89 \\ 
        GUPNet \cite{lu2021gupnet}$\dagger$   & ICCV 2021 & -     & 20.11 & \blue{14.20} & 11.77 & - & - & -  \\ \hline
        
        MonoDTR (Ours)                        & -         & 37    & \red{21.99} & \red{15.39} &  \red{12.73} & \blue{28.59} & \red{20.38} & 17.14 \\ 
    \end{tabular}
    \vspace{-2mm}
    \caption{\textbf{Detection performance of Car category on the KITTI test set.} 
    The best and second best results are highlighted in \red{red} and \blue{blue}, respectively.
    $\dagger$ indicates the results are reported in their papers.
    }
    \label{tab:kitti_test_car}
\end{table*}

\begin{table*}[ht]
    \centering
    \small
    \begin{tabular}{l|ccc|ccc|ccc|ccc}  
        \hline 
        \multicolumn{1}{l|}{\multirow{2}{*}{Method}} &
        \multicolumn{3}{c|}{\apthreeD@IoU=0.7} &  \multicolumn{3}{c|}{\apBev@IoU=0.7} &  \multicolumn{3}{c|}{\apthreeD@IoU=0.5} & \multicolumn{3}{c}{\apBev@IoU=0.5} \\ \cline{2-13}  
        & Easy  & Mod. & Hard & Easy & Mod.  & Hard & Easy & Mod.  & Hard & Easy & Mod.  & Hard  \\ \hline \hline
        M3D-RPN \cite{brazil2019m3drpn}         & 14.53 & 11.07 & 8.65  & 20.85 & 15.62 & 11.88 & 48.53 & 35.94 & 28.59 & 53.35 & 39.60 & 31.76 \\
        MonoPair \cite{chen2020monopair}        & 16.28 & 12.30 & 10.42 & 24.12 & 18.17 & 15.76 & 55.38 & 42.39 & 37.99 & 61.06 & 47.63 & 41.92 \\
        MonoDLE \cite{Ma2021monodle}            & 17.45 & 13.66 & 11.68 & 24.97 & 19.33 & 17.01 & 55.41 & 43.42 & 37.81 & 60.73 & 46.87 & 41.89 \\
        Kinematic3D \cite{brazil2020kinematic}  & 19.76 & 14.10 & 10.47 & 27.83 & 19.72 & 15.10 & 55.44 & 39.47 & 31.26 & 61.79 & 44.68 & 34.56 \\
        GrooMeD-NMS \cite{kumar2021groomed}     & 19.67 & 14.32 & 11.27 & 27.38 & 19.75 & 15.92 & 55.62 & 41.07 & 32.89 &61.83 & 44.98 & 36.29 \\

        MonoRUn \cite{chen2021monorun}          & 20.02 & 14.65 & 12.61 &  -    &   -   &   -   & 59.71 & 43.39 & 38.44 & - & - & - \\ 
        CaDDN \cite{Reading2021CaDDN}           & 23.57 & 16.31 & 13.84 & - & - & - & - & - & - & - & - & - \\
        GUPNet \cite{luo2021m3dssd}             & 22.76 & 16.46 & 13.72 &31.07 &22.94 &19.75 &57.62 &42.33 &37.59 &61.78 &47.06 &40.88 \\
        MonoFlex \cite{Zhang2021MonoFlex}       & 23.64 & 17.51 & 14.83 & - & - & - & - & - & - & - & - & - \\ 
        \hline
        MonoDTR (Ours)                & \textbf{24.52} & \textbf{18.57} & \textbf{15.51} & \textbf{33.33} & \textbf{25.35} & \textbf{21.68} & \textbf{64.03} & \textbf{47.32} & \textbf{42.20} & \textbf{69.04} & \textbf{52.47} & \textbf{45.90} \\
    \end{tabular}
    \vspace{-2mm}
    \caption{\textbf{Detection performance of Car category on the KITTI validation set.} We utilize \textbf{bold} to highlight the best results.}
    \label{tab:kitti_val}
\end{table*}

\smallskip\noindent{\bf Implementation details.}
We use Adam optimizer to train our network for 120 epochs with batch size 4. 
The learning rate starts at 0.0001 and decays with a cosine annealing schedule. 
We apply 48 anchors on each pixel of the feature map with 3 aspect ratios of $\{0.5, 1.0, 1.5\}$, and 12 scales in height following the exponential function $24 \times 2^{i/4}, i = \{0,...,15\}$. For 3D anchor parameters, we calculate the mean and variance statistics of 3D ground truth in the training dataset as prior statistical knowledge of each anchor. Following \cite{liu2021ground}, we crop the top 100 pixels of each image to reduce inference time, and all images are resized to 288 $\times$ 1280. 
In the training stage, we apply random horizontal mirroring as data augmentation.
In the inference stage, we drop the predictions with a confidence score lower than 0.75 and adopt Non-Maximum Suppression (NMS) with IoU 0.4 to reduce redundancy.

\subsection{Main Results}
\noindent{\bf Results of the Car category on the KITTI test set.}
As shown in Table \ref{tab:kitti_test_car}, we compare our MonoDTR with several state-of-the-art monocular 3D object detection methods on the KITTI test set.
It can be observed that our approach achieves better performance than other methods in terms of the moderate level of the two tasks, which is the most important metric in the benchmark.
Furthermore, it is worth noting that our approach outperforms other depth-assisted methods by large margins.
For instance,
compared to top three depth-assist methods, DFRNet \cite{Zou2021dfr}, CaDDN \cite{Reading2021CaDDN} and  DDMP-3D \cite{wang2021ddmp},
our MonoDTR obtains \textbf{2.59/1.76/2.38}, \textbf{2.82/1.98/1.27} and \textbf{2.28/2.61/2.93} improvements in \apthreeD~at IoU threshold 0.7 on three settings, which indicates the effectiveness of the proposed depth-aware modules.

\smallskip{\noindent{\bf Results of the Car category on the KITTI validation set.}}
We also conduct experiments on the KITTI validation dataset under different IoU thresholds and tasks as listed in Table \ref{tab:kitti_val}.
Our approach obtains superior performance over several image-only methods, benefiting from the auxiliary depth supervision.
Specifically, compared with GUPNet \cite{lu2021gupnet}, our method achieves significant improvements of \textbf{6.41/4.99/4.61} in \apthreeD~and \textbf{7.26/5.41/5.02} in \apBev~at IoU threshold 0.5 on the easy, moderate, and hard settings.

\smallskip{\noindent{\bf Results of Pedestrians and Cyclists categories on the KITTI test set.}}
We further present the performance of pedestrians and cyclists categories in Table \ref{tab:kitti_ped_cyc}. 
Detecting these two categories is more challenging than cars due to their smaller size and non-rigid body, making it difficult to precisely locate the position.
Overall, our model significantly outperforms all methods on pedestrian category with a considerable margin. For the cyclist 3D detection, we also achieve competitive results to CaDDN \cite{Reading2021CaDDN} and obtain better performance than other methods.

\setlength{\tabcolsep}{0.015\linewidth}{
\begin{table}[t]
    \small
    \centering
    \begin{tabular}{l|ccc|ccc}  
        \hline 
        \multicolumn{1}{l|}{\multirow{2}{*}{Method}} &
        \multicolumn{3}{c|}{\apthreeD (Ped.)} &  \multicolumn{3}{c}{\apthreeD (Cyc.)}  \\ \cline{2-7} 
        & Easy  & Mod. & Hard & Easy & Mod.  & Hard   \\ \hline \hline
        MonoDLE \cite{Ma2021monodle}        & 9.64  & 6.55  & 5.44 & 4.59 & 2.66 & 2.45  \\
        MonoPair \cite{chen2020monopair}    & 10.02 & 6.68  & 5.53 & 3.79 & 2.12 & 1.83  \\
        MonoFlex \cite{Zhang2021MonoFlex}   & 9.43  & 6.31  & 5.26 & 4.17 & 2.35 & 2.04  \\ \hline 
        D4LCN \cite{ding2020d4lcn}          & 4.55  & 3.42  & 2.83 & 2.45 & 1.67 & 1.36    \\
        DDMP3D \cite{wang2021ddmp}          & 4.93  & 3.55  & 3.01 & 4.18 & 2.50 & 2.32  \\
        CADDN \cite{Reading2021CaDDN}       & 12.87 & 8.14  & 6.76 & \textbf{7.00} & \textbf{3.41} & \textbf{3.30}  \\     
        \hline
        MonoDTR (Ours)                          & \textbf{15.33} &\textbf{10.18}  & \textbf{8.61} & 5.05 & 3.27 & 3.19  \\ 
    \end{tabular}
    \vspace{-2mm}
    \caption{\textbf{Detection performance of Pedestrian and Cyclist categories on the KITTI test set} at 0.5 IoU threshold. We utilize \textbf{bold} to highlight the best results.}
    \label{tab:kitti_ped_cyc}
\end{table}
}

\smallskip{\noindent{\bf Running time analysis.}}
We measure the average running time for processing the whole validation set with batch size 1 on a single Nvidia Tesla v100 GPU.
As shown in Table \ref{tab:kitti_test_car}, our model can achieve real-time performance at 27 FPS, which confirms the efficiency of our approach.
Compared with state-of-the-art depth-assisted methods, our MonoDTR runs 17$\times$ and 4.8$\times$ faster than CaDDN \cite{Reading2021CaDDN} and DDMP-3D \cite{wang2021ddmp}, respectively.
The main reasons can be summarized as follows:
(1) CaDDN \cite{Reading2021CaDDN} builds the bird's eye view representation from predicted depth maps to perform 3D detection, which applies more complicated architecture to generate precise depth predictions, leading to slow speed.
(2) Fusion-based methods \cite{ding2020d4lcn, wang2021ddmp} often utilize two separate backbones for extracting features of image and depth, which is time-consuming.
Note that the depth estimator also takes additional inference time, which is not included in Table \ref{tab:kitti_test_car}.
On the contrary, our model learns depth-aware features through the lightweight DFE module with auxiliary supervision, 
which reduces running time significantly.

\subsection{Ablation Study}
\label{sec:ablation}
\smallskip{\noindent{\bf Effectiveness of each proposed components.}}
In Table \ref{tab:abl_arch}, we conduct an ablation study to analyze the effectiveness of the proposed components:
(a) Baseline: only using context-aware features for 3D object detection, \ie, without all proposed depth-aware modules.
(b) Replacing depth-aware features with object query \cite{Nicolas2020detr} in the transformer, \ie baseline + DETR-like transformer.
(c) Replacing depth-aware features with features extracted from depth images generated by DORN \cite{FuCVPR18-DORN}.
(d) Integrating context- and depth-aware features with the convolutional concatenate operation.
(e) Full model without depth prototype enhanced feature $\mathbf{F'}$.
(f) MonoDTR (full model).

\begin{figure}[t]
    \centering
    \includegraphics[width=0.882\columnwidth]{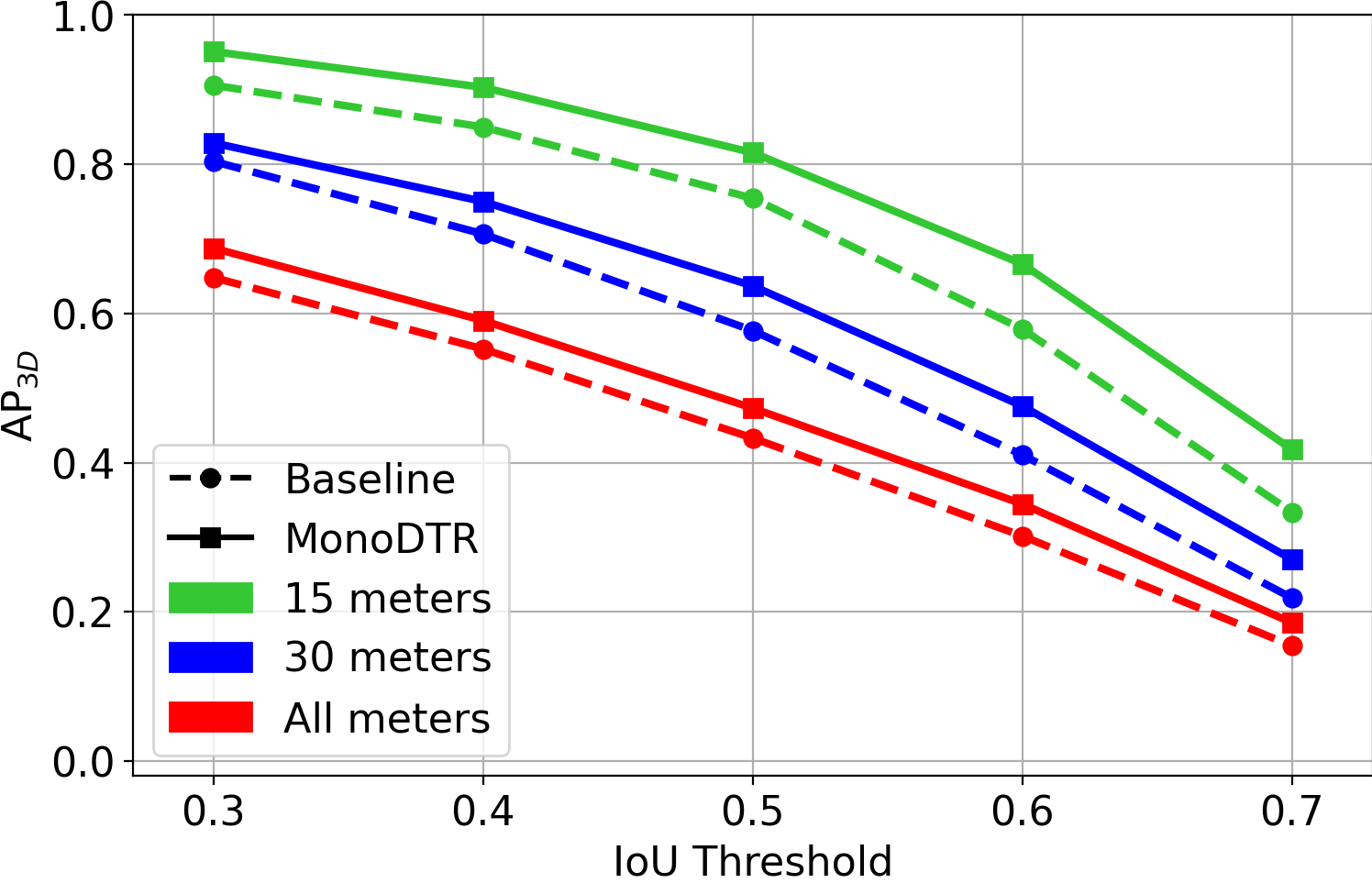}
    \vspace{-2.3mm}
    \caption{\textbf{Comparison of AP with different object depth ranges and IoU thresholds between baseline and MonoDTR} on the KITTI validation set for Car category. Best viewed in color.}
    \label{fig:ap3d_vs_depth}
\end{figure}

Firstly, we can observe from (b$\rightarrow$f) that utilizing depth-aware features to replace the object query in the transformer can provide meaningful depth hints and improve the performance.
Besides, compared to our end-to-end training framework (f), simply utilizing depth priors from the pretrained depth estimator (c) leads to worse results.
Next, we demonstrate that applying our depth-aware transformer (DTR) module (f) can more effectively integrate context- and depth-aware features than simple convolutional concatenation (d).
Furthermore, utilizing our proposed depth prototype enhancement module can boost the performance (e$\rightarrow$f).
Finally, by applying all the designed modules, our full model (f) achieves significant improvement compared to the baseline (a). 
Also, an in-depth analysis in Figure \ref{fig:ap3d_vs_depth} suggests that our method surpasses the baseline under different IoU thresholds and object depths.
These results prove the effectiveness of our depth-aware modules.

\setlength{\tabcolsep}{0.011\linewidth}{
\begin{table}[t]
    \small
    \centering
    \begin{tabular}{c|l|ccc}  
        \hline  
        \multicolumn{1}{l|}{\multirow{2}{*}{}}
        &
        \multicolumn{1}{l|}{\multirow{2}{*}{Ablation}}
        & \multicolumn{3}{c}{\apthreeD@IoU=0.7}  \\ \cline{3-5} 
        & & Easy  & Mod.   & Hard    \\ 
        \hline \hline
        (a) & Baseline &  19.35&  15.47&  12.83    \\
        (b) & depth-aware feature $\rightarrow$ object query & 20.09 & 16.10 & 14.07 \\
        (c) & depth-aware feature $\rightarrow$ DORN \cite{FuCVPR18-DORN} & 24.08 & 17.10 & 14.02\\
        (d) & DTR $\rightarrow$ concat. operation  & 23.39 & 17.65 & 14.82 \\
        (e) & w/o depth prototype enhancement & 23.72 & 18.22 & 15.36  \\  
        (f) & MonoDTR (full model) & \textbf{24.52} & \textbf{18.57} & \textbf{15.51}\\ \hline
    \end{tabular}
    \vspace{-2mm}
    \caption{\textbf{Analysis of different components of our approach} on the KITTI validation set for Car category. 
    }
    \label{tab:abl_arch}
\end{table}
}

\begin{figure*}[th]
\centering
\includegraphics[width=0.98\textwidth]{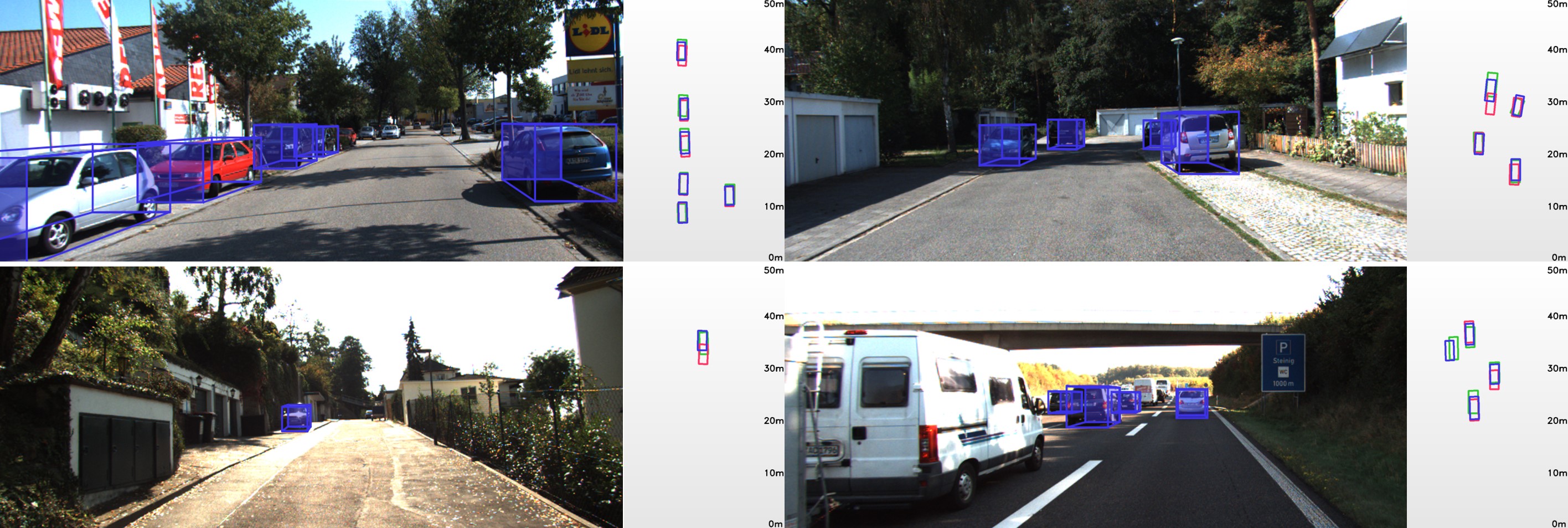}
\caption{\textbf{Qualitative examples on the KITTI validation set.} We provide the predictions  on the image view (left) and bird eye view (right). The \textcolor{plotpurple}{purple} boxes in the image and BEV plane represent the predictions from MonoDTR. The \textcolor{plotgreen}{green} and \textcolor{plotred}{pink} boxes on BEV are the ground truth and the predictions from baseline (without depth-aware modules), respectively. Best viewed in color and zoomed in.
}
\label{fig:kitti_vis}
\vspace{-6pt}
\end{figure*}

\smallskip{\noindent{\bf Comparison with different positional encodings.}}
We investigate the effectiveness of the proposed depth positional encoding (DPE) in Table \ref{tab:abl_pe}. 
Compared with several commonly used positional encodings, including absolute positional encoding (APE) \cite{dosovitskiy2020vit}, conditional positional encoding (CPE) \cite{chu2021conditional}, sinusoidal positional encoding \cite{vaswani2017SA}, and without using positional encoding (No PE), our proposed DPE achieves better performance on KITTI validation set. 
We believe that encoding the depth-aware cues is more effective for learning the position representation of 3D tasks than pixel-level encodings.

\setlength{\tabcolsep}{0.013\linewidth}{
\begin{table}[t]
    \small
    \centering
    \begin{tabular}{l|ccc|ccc}  
        \hline   
        \multicolumn{1}{l|}{\multirow{2}{*}{Positional Enc.}}& \multicolumn{3}{c|}{\apthreeD@IoU=0.7} &  \multicolumn{3}{c}{\apBev@IoU=0.7}  \\ 
        \cline{2-7}   
        & Easy  & Mod. & Hard & Easy & Mod.  & Hard   \\ \hline \hline
        No PE                                   & 23.65 &  17.76 & 15.05 & 31.33 & 24.02 & 20.83 \\
        Sinusoidal \cite{vaswani2017SA}         & 22.73 & 17.63 & 14.74 & 31.78 & 24.40 & 20.97   \\
        APE  \cite{dosovitskiy2020vit}      & 23.85 & 17.55 & 14.59 & 32.52 & 23.47 & 19.92   \\
        CPE \cite{chu2021conditional}           &   24.34 & 18.04 & 15.14 & 33.01 & 24.69 & 20.48 \\\hline
        DPE (Ours)      & \textbf{24.52} & \textbf{18.57} & \textbf{15.51} & \textbf{33.33} & \textbf{25.35} & \textbf{21.68} \\
    \end{tabular}
    \vspace{-2mm}
    \caption{\textbf{
    Comparison of different positional encoding mechanisms} on the KITTI validation set for Car category.}
    \label{tab:abl_pe}
\end{table}
}

\smallskip{\noindent{\bf Plugging into the existing image-only methods.}}
Our proposed approach is flexible to extend to existing image-only 3D object detectors to improve the depth reasoning capability. We respectively plug our depth-aware modules into three popular monocular 3D object detectors: M3D-RPN \cite{brazil2019m3drpn}, GAC \cite{liu2021ground}, and MonoDLE \cite{Ma2021monodle}, based on their official codes\footnote{\url{https://github.com/garrickbrazil/M3D-RPN}}\footnote{\url{https://github.com/Owen-Liuyuxuan/visualDet3D}}\footnote{\url{https://github.com/xinzhuma/monodle}}. 
In practice, we take the features from the above models (before the detection head) as the initial features, and utilize our proposed modules (DFE, DTR, and DPE modules) to generate final integrated features, followed by their original detection head to detect 3D objects.
As shown in Table \ref{tab:abl_equip}, with the aid of our proposed depth-aware modules, these detectors achieve further improvements on the KITTI validation set, which demonstrates the flexibility and efficiency of our approach.

\subsection{Qualitative Results}
We provide the qualitative examples on the KITTI validation set in Figure \ref{fig:kitti_vis}.
Compared with the baseline model without the aid of depth-aware modules, the predictions from MonoDTR are much closer to the ground truth.
It shows that the proposed depth-aware modules can help to locate the object precisely.
More qualitative results are included in the supplementary material.

\setlength{\tabcolsep}{0.011\linewidth}{
\begin{table}[t]
    \small
    \centering
    \begin{tabular}{l|ccc|ccc}  
        \hline  
        \multicolumn{1}{l|}{\multirow{2}{*}{Method}}
        & \multicolumn{3}{c|}{\apthreeD@IoU=0.7} &  \multicolumn{3}{c}{\apBev@IoU=0.7}  \\ \cline{2-7} 
        & Easy  & Mod. & Hard & Easy & Mod.  & Hard   \\ \hline \hline
        M3D-RPN \cite{brazil2019m3drpn}  & 14.53 & 11.07 &  8.65 & 20.85 & 15.62 & 11.88 \\
        M3D-RPN + \textbf{Ours} & \textbf{20.96} & \textbf{16.44} & \textbf{14.63} & \textbf{25.24} & \textbf{20.52} & \textbf{17.43} \\ 
        \cellcolor{TableBlue}Improvement & \cellcolor{TableBlue}+6.43 & \cellcolor{TableBlue}+5.37 & \cellcolor{TableBlue}+5.98 & \cellcolor{TableBlue}+4.39 & \cellcolor{TableBlue}+4.90 & \cellcolor{TableBlue}+5.55  \\ \hline

        GAC \cite{liu2021ground}* & 21.58 & 15.17 & 11.35 & 28.62 & 19.99 & 15.42   \\
        GAC + \textbf{Ours}& \textbf{24.30} & \textbf{17.28} & \textbf{13.35} & \textbf{33.02} & \textbf{23.06} & \textbf{18.22}   \\ 
        \cellcolor{TableBlue}Improvement & \cellcolor{TableBlue}+2.72 & \cellcolor{TableBlue}+2.11 & \cellcolor{TableBlue}+2.00 & \cellcolor{TableBlue}+4.40 & \cellcolor{TableBlue}+3.07 & \cellcolor{TableBlue}+2.80  \\ \hline
        
         MonoDLE \cite{Ma2021monodle} & 17.45 & 13.66 & 11.68 &24.97 & 19.33 &17.01 \\

         MonoDLE + \textbf{Ours} & \textbf{18.68} & \textbf{15.69} & \textbf{13.41} & \textbf{26.67} & \textbf{21.40} & \textbf{18.67} \\
        \cellcolor{TableBlue}Improvement & \cellcolor{TableBlue}+1.23 & \cellcolor{TableBlue}+2.03 & \cellcolor{TableBlue}+1.73 & \cellcolor{TableBlue}+1.70 & \cellcolor{TableBlue}+2.07 & \cellcolor{TableBlue}+1.66  \\ \hline

    \end{tabular}
    \vspace{-2mm}
    \caption{\textbf{Extension on existing image-only monocular 3D object detectors.} 
    * indicates that we retrained without using extra right images. See details in Section \ref{sec:ablation}.}
    \label{tab:abl_equip}
\end{table}
}

\section{Conclusion}

In this paper, we propose a depth-aware transformer network for monocular 3D object detection. The proposed lightweight DFE module implicitly learns depth-aware features in an end-to-end fashion to avoid obtaining inaccurate depth priors and high computational cost from an off-the-shelf depth estimator. We also introduce the depth-aware transformer to globally integrate context- and depth-aware features, while the novel depth positional encoding (DPE) is designed to inject depth hints into the transformer.
Comprehensive experiments on the KITTI dataset validate that our model achieves real-time detection and outperforms previous state-of-the-art monocular-based methods.
\vspace{-5.5mm}
\paragraph{Acknowledgements.}
This work was supported in part by the Ministry of Science  and  Technology, Taiwan, under Grant MOST 110-2634-F-002-051, Qualcomm Technologies, Inc., and Mobile Drive Technology Co., Ltd (MobileDrive). We are grateful to the National Center for High-performance Computing.

{\small
\bibliographystyle{ieee_fullname}
\bibliography{egbib}
}
\clearpage

\appendix
\noindent{\bf\Large Supplementary Material}
\section{Depth-Aware Transformer}
\begin{figure}[h]
    \centering
    \includegraphics[width=0.97\columnwidth]{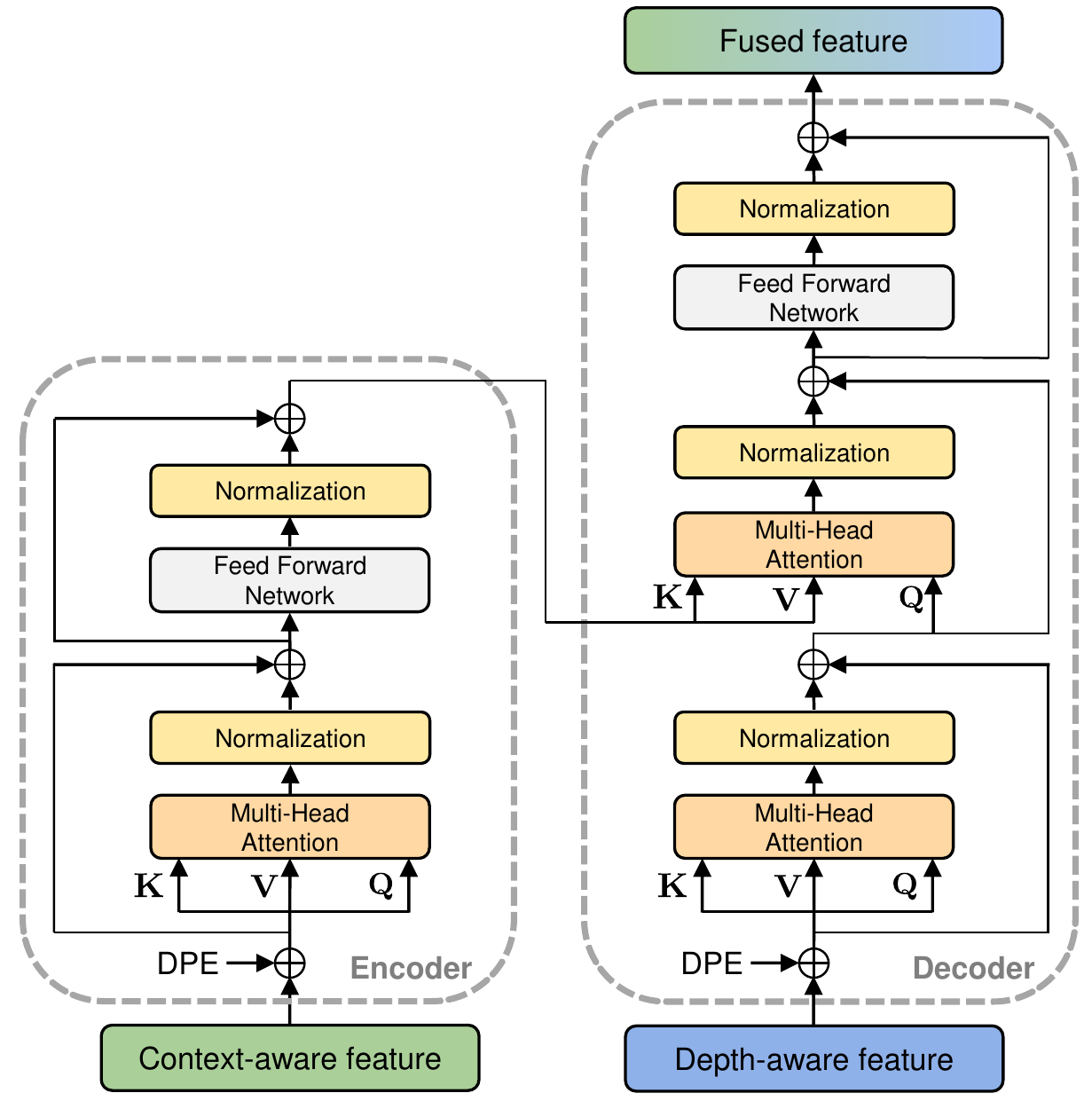} %
    \caption{\textbf{The architecture of depth-aware transformer (DTR).} DPE is the depth positional encoding proposed in the main paper.
    }
    \vspace{-5pt}
    \label{fig:arch_dtr}
\end{figure}

\noindent{\bf Transformer architecture.}
The detailed architecture of depth-aware transformer (DTR) is shown in Figure \ref{fig:arch_dtr}. The encoder aims to generate the encoded context-aware features, while the decoder produces the fused feature from context- and depth-aware features through the multiple self-attention layers. Besides, we supplement two features with the proposed depth positional encoding (DPE) before passing them to the transformer, enabling better 3D reasoning.

\smallskip{\noindent{\textbf{Effectiveness of linear attention.}}}
Table \ref{tab:abl_sa} shows the results of different self-attention layers on the KITTI dataset, where we can observe that applying linear attention \cite{katharopoulos2020rnn} can achieve almost 4 $\times$ faster than
vanilla self-attention \cite{vaswani2017SA} with comparable performance. Thus, we adopt the linear attention \cite{katharopoulos2020rnn} in our transformers for real-time applications.

\setlength{\tabcolsep}{0.009\linewidth}{
\begin{table}[htpb]
    \small
    \centering
    \begin{tabular}{l|c|ccc|ccc}  
        \hline 
        \multicolumn{1}{l|}{\multirow{2}{*}{Attention}} & 
        \multicolumn{1}{c|}{\multirow{2}{*}{Time}}  &
        \multicolumn{3}{c|}{\apthreeD@IoU=0.7} &  \multicolumn{3}{c}{\apBev@IoU=0.7} \\ \cline{3-8}  &   
        & Easy  & Mod. & Hard  & Easy  & Mod. & Hard   \\ \hline \hline
        vanilla SA  \cite{vaswani2017SA}             &      136 ms& 24.38 & 18.39 & 16.35 & 31.57 & 24.51 & 21.40   \\
        linear SA \cite{katharopoulos2020rnn}     &      37 ms & 24.52 & 18.57 & 15.51 & 33.33 & 25.35 & 21.68   \\ \hline
    \end{tabular}
    \vspace{-2mm}
    \caption{\textbf{Comparison of different self-attention mechanisms} on the KITTI validation set for Car category. We follow the same setting and device as in the main paper for running time measurement.
    Note that 'SA' is the multi-head self-attention. The metric is \apForty.}
    \label{tab:abl_sa}
\end{table}
}

\section{Auxiliary Depth Supervision}
\noindent{\textbf{Depth ground truth generation.}} 
We project the LiDAR signals into the image plane to generate the sparse ground truth depth map. Then we apply linear-increasing discretization (LID) \cite{tang2020center3d} method to convert continuous depth $d$ to discretized depth bins. The LID is defined as follows:
\vspace{-2mm}
\begin{align} 
\label{eq:disc}
     d &=  d_{\min} + \frac{d_{\max}-d_{\min}}{D(D + 1)} \cdot i (i + 1),~ i=\{1,..., D\},
\end{align}
where $i$ is the depth bin index. The number of depth bins $D$ is set as 96, and the range of depth [$d_{\min}, d_{\max}$] is set as [1, 80]. Note that the pixels with the depth value outside the range will be marked as invalid and not used for optimization during training.

\smallskip{\noindent{\textbf{Different discretization methods.}}}
In Table \ref{tab:abl_depth}, we investigate the effectiveness of different discretization methods for depth auxiliary supervision. 
In addition to the LID method, the continuous depth can be discretized using uniform discretization (UD) with fixed bin size: $\frac{d_{\max}-d_{\min}}{D}$, or spacing-increasing discretization (SID) \cite{FuCVPR18-DORN} with the increasing bin size in the log space.
It can be observed that using the LID strategy can achieve better performance, so we apply it as our discretization method.

\setlength{\tabcolsep}{0.010\linewidth}{
\begin{table}[h]
    \vspace{3mm}
    \small
    \centering
    \begin{tabular}{c|ccc|ccc}  
        \hline
        \multicolumn{1}{c|}{\multirow{2}{*}{Disc. Method}}
        & \multicolumn{3}{c|}{\apthreeD@IoU=0.7}   
        & \multicolumn{3}{c}{\apBev@IoU=0.7} 
        \\ \cline{2-7} 
        & Easy  & Mod. & Hard & Easy  & Mod. & Hard   \\ \hline \hline
        UD & 23.22 & 17.67 & 14.80 & 31.75 & 24.32 & 20.08 \\
        SID& 23.89 & 18.10 & 15.22 & 32.19 & 24.76 & 21.36  \\
        LID& \textbf{24.52} & \textbf{18.57} & \textbf{15.51} & \textbf{33.33} & \textbf{25.35} & \textbf{21.68}   \\ \hline
    \end{tabular}
    \vspace{-2mm}
    \caption{\textbf{Comparison of different discretization methods for auxiliary depth supervision} on the KITTI validation set for Car category. The metric is \apForty.}
    \label{tab:abl_depth}
\end{table}
}

\begin{figure*}[t]
\centering
\includegraphics[width=0.99\textwidth]{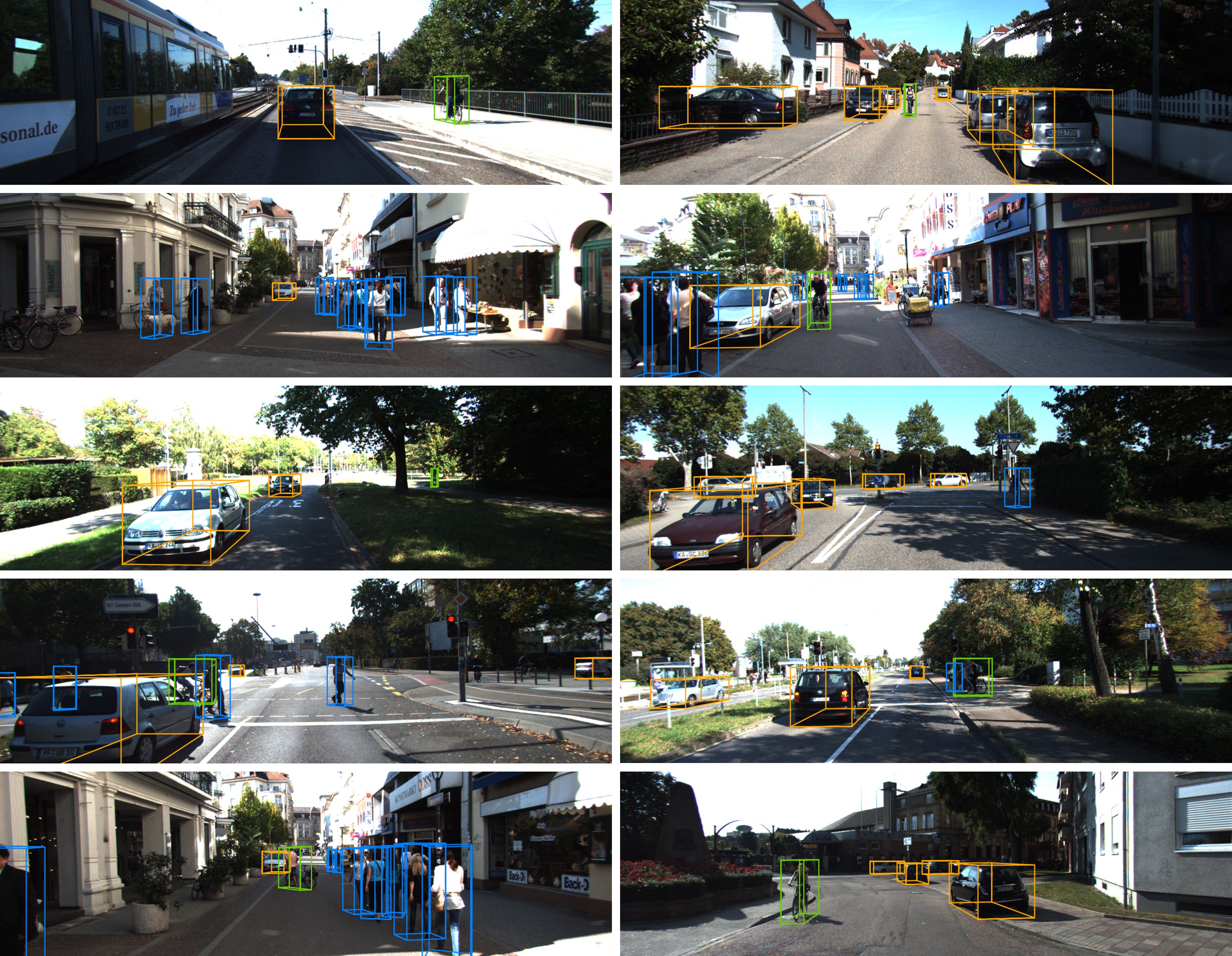}
\caption{\textbf{Qualitative results on the KITTI validation set for multi-class 3D object detection.} We utilize \textcolor[RGB]{255,128,0}{orange}, \textcolor[RGB]{0,128,255}{blue}, and 
\textcolor[RGB]{102,204,0}{green} colors to indicate car, pedestrian, and cyclist categories, respectively.
}
\label{fig:kitti_vis_multi}
\vspace{-6pt}
\end{figure*}

\section{Results on nuScenes Dataset}
Table \ref{tab:nuscene_val} shows the experimental results of deploying our proposed approach on nuScenes \cite{nuscenes2019} val set. 
Under the same configurations (\eg, backbone and training schedule), our model achieves better performance than two 3D object detection baselines (FCOS3D \cite{wang2021fcos3d}, and PGD \cite{wang2021pgd}), which demonstrates the effectiveness of our approach.

\setlength{\tabcolsep}{0.017\linewidth}{
\begin{table}[htpb]
    \small
    \centering
    \begin{tabular}{l|ccccc}  
        \hline 
        Method & NDS $\uparrow$ & mAP $\uparrow$  \\ \hline \hline
        FCOS3D \cite{wang2021fcos3d} & 37.7 & 29.8   \\
        PGD \cite{wang2021pgd} &  39.3 & 31.7    \\ \hline
        Ours & \textbf{40.1} & \textbf{33.8}   \\
    \end{tabular}
    \vspace{-3mm}
    \caption{\textbf{Detection performance on nuScenes val set.}
    We build our approach based on FCOS3D \cite{wang2021fcos3d}. The experiments are conducted under the same training settings (trained for 12 epochs).
    The results of baselines are taken from MMDetection3D \cite{mmdet3d2020}.} 
    \label{tab:nuscene_val}
\end{table}
}

\section{Qualitative Visualization}

\begin{figure*}[th]
\centering
\includegraphics[width=0.99\textwidth]{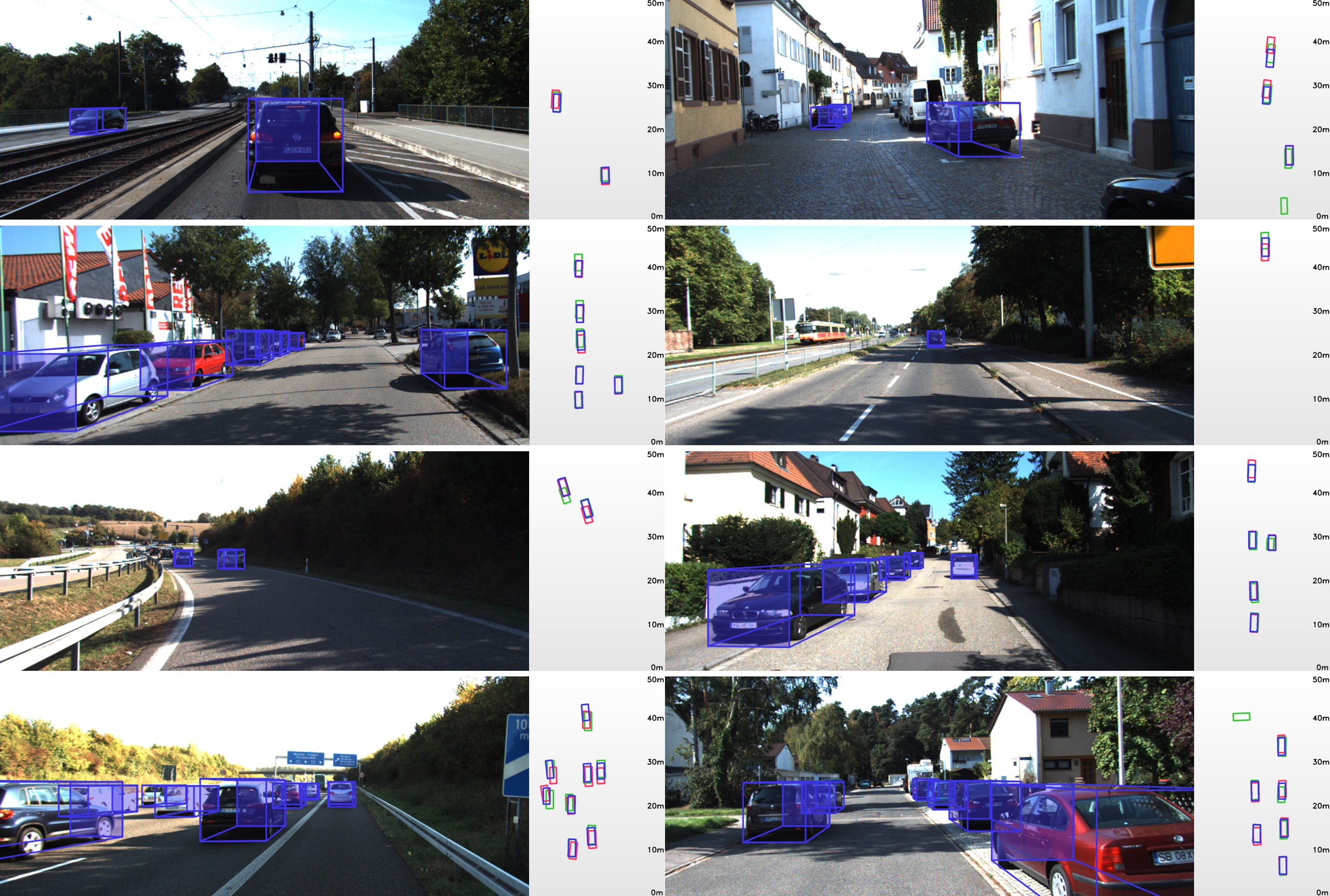}
\caption{\textbf{Qualitative comparison on the KITTI validation set} for the car category. The \textcolor{plotpurple}{purple} boxes in the image and BEV plane represent the predictions from MonoDTR. The \textcolor{plotgreen}{green} and \textcolor{plotred}{pink} boxes on BEV are the ground truth and the predictions from baseline (our full model without proposed depth-aware modules), respectively. Best viewed in color and zoomed in.
}
\label{fig:kitti_vis_supp}
\vspace{-6pt}
\end{figure*}

\begin{figure*}[th]
\centering
\includegraphics[width=0.9\textwidth]{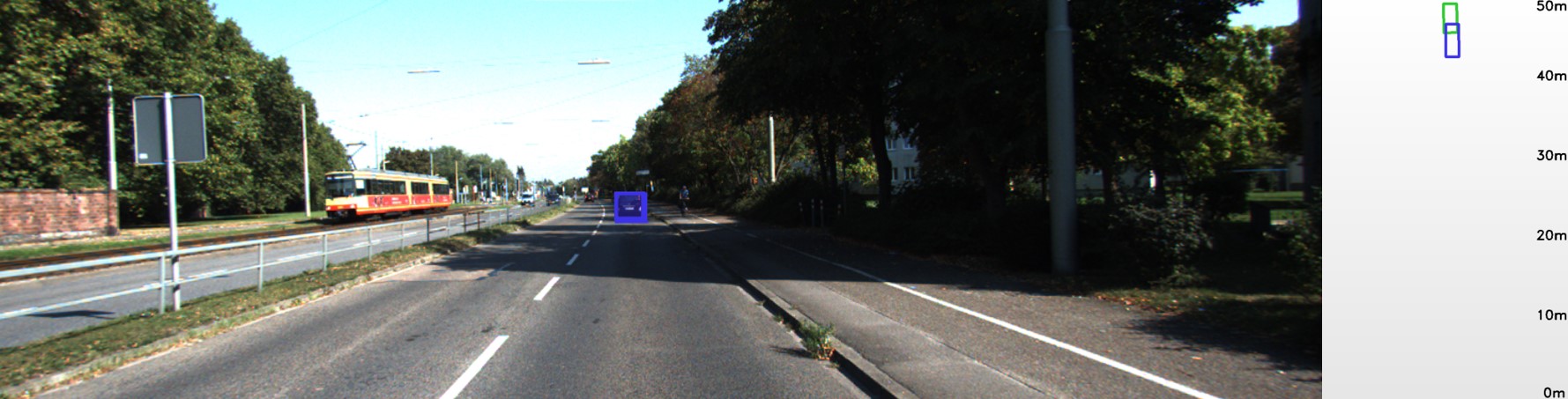}
\caption{\textbf{Failure case.} The \textcolor{plotpurple}{purple} and \textcolor{plotgreen}{green} boxes represent the predictions from MonoDTR and ground truth, respectively. The failure case is caused by the inaccurate object center depth estimation.}
\label{fig:kitti_fail}
\vspace{-6pt}
\end{figure*}

\noindent{\textbf{More visualization results.} 
In Figure \ref{fig:kitti_vis_multi}, we provide some qualitative results on the KITTI dataset for multiple-category predictions. 
In Figure \ref{fig:kitti_vis_supp}, we show the qualitative comparisons of the baseline (without proposed depth-aware modules) and our MonoDTR (full model). It can be observed that our MonoDTR can generate higher quality bounding boxes benefit from the aid of depth cues.
}

\smallskip{\noindent{\textbf{Failure case.}} 
We show a representative failure case in Figure \ref{fig:kitti_fail}. The lower-quality 3D bounding box is caused by the inaccurately predicted object depth, which is typical in most monocular 3D object detection tasks.}

\section{Broader Impacts}
Our work aims to develop the monocular 3D object detection approach for autonomous driving. The proposed model may generate inaccurate object depth prediction, leading to incorrect downstream decision-making and potential traffic accidents. 
Furthermore, we provide a new perspective of leveraging learned depth-aware features to assist monocular 3D object detection. Although considerable progress has been made with our proposed lightweight depth-aware feature extraction module, we believe it is worth further exploring how to learn depth-aware features to effectively improve detection performance.

\end{document}